\documentclass[10pt,twocolumn,letterpaper]{article}

\usepackage[pagenumbers]{cvpr} 

\usepackage{cite}
\usepackage{amsmath,amssymb,amsfonts}
\usepackage{amsthm,amsmath,amssymb}
\usepackage{mathrsfs}
\usepackage{textcomp}
\usepackage{xcolor}
\usepackage{bm}
\usepackage{amsfonts}
\usepackage{multirow}
\usepackage{verbatim}
\usepackage{tabularx}

\usepackage{graphicx}
\usepackage{subcaption}

\usepackage{booktabs}
\usepackage[ruled,linesnumbered]{algorithm2e}

\usepackage[pagebackref,breaklinks,colorlinks]{hyperref}

\usepackage[capitalize]{cleveref}
\crefname{section}{Sec.}{Secs.}
\Crefname{section}{Section}{Sections}
\Crefname{table}{Table}{Tables}
\crefname{table}{Tab.}{Tabs.}

\def\cvprPaperID{1372} 
\def\confName{CVPR}
\def\confYear{2022}

\begin{document}

\title{CP-ViT: Cascade Vision Transformer Pruning via Progressive Sparsity Prediction}

\author{Zhuoran Song*, Yihong Xu*, Zhezhi He, Li Jiang, Naifeng Jing, and Xiaoyao Liang\\
Shanghai Jiao Tong University, China\\
{\tt\small songzhuoran@sjtu.edu.cn}
}
\maketitle

\begin{abstract}
   Vision transformer (ViT) has achieved competitive accuracy on a variety of computer vision applications, but its computational cost impedes the deployment on resource-limited mobile devices. 
   We explore the sparsity in ViT and observe that informative patches and heads are sufficient for accurate image recognition. 
   In this paper, we propose a cascade pruning framework named CP-ViT by predicting sparsity in ViT models progressively and dynamically to reduce computational redundancy while minimizing the accuracy loss. Specifically, we define the cumulative score to reserve the informative patches and heads across the ViT model for better accuracy. We also propose the dynamic pruning ratio adjustment technique based on layer-aware attention range. CP-ViT has great general applicability for practical deployment, which can be applied to a wide range of ViT models and can achieve superior accuracy with or without fine-tuning. 
   Extensive experiments on ImageNet, CIFAR-10, and CIFAR-100 with various pre-trained models have demonstrated the effectiveness and efficiency of CP-ViT. By progressively pruning 50\% patches, our CP-ViT method reduces over 40\% FLOPs while maintaining accuracy loss within 1\%.
\end{abstract}



\section{Introduction}
\label{sec:intro}

Recently, Transformers have demonstrated great successes in computer vision, such as image classification~\cite{touvron2021training,wu2020lite,han2021transformer,szegedy2017inception,rawat2017deep}, object detection~\cite{zhu2020deformable,chi2020relationnet++,dai2021up,yang2021uncertainty}, semantic segmentation~\cite{badrinarayanan2017segnet,song2020vr}, and action recognition~\cite{plizzari2021spatial,li2021trear,girdhar2019video}. 
Thanks to the self-attention based architectures, Vision Transformers~(ViT)~\cite{dosovitskiy2020image} outperforms the classical Convolutional Neural Networks~(CNN)~\cite{shin2016deep} which achieves the state-of-the-art results.


\begin{figure}[h]
\centering\vspace{-10pt}
\includegraphics[width=\linewidth]{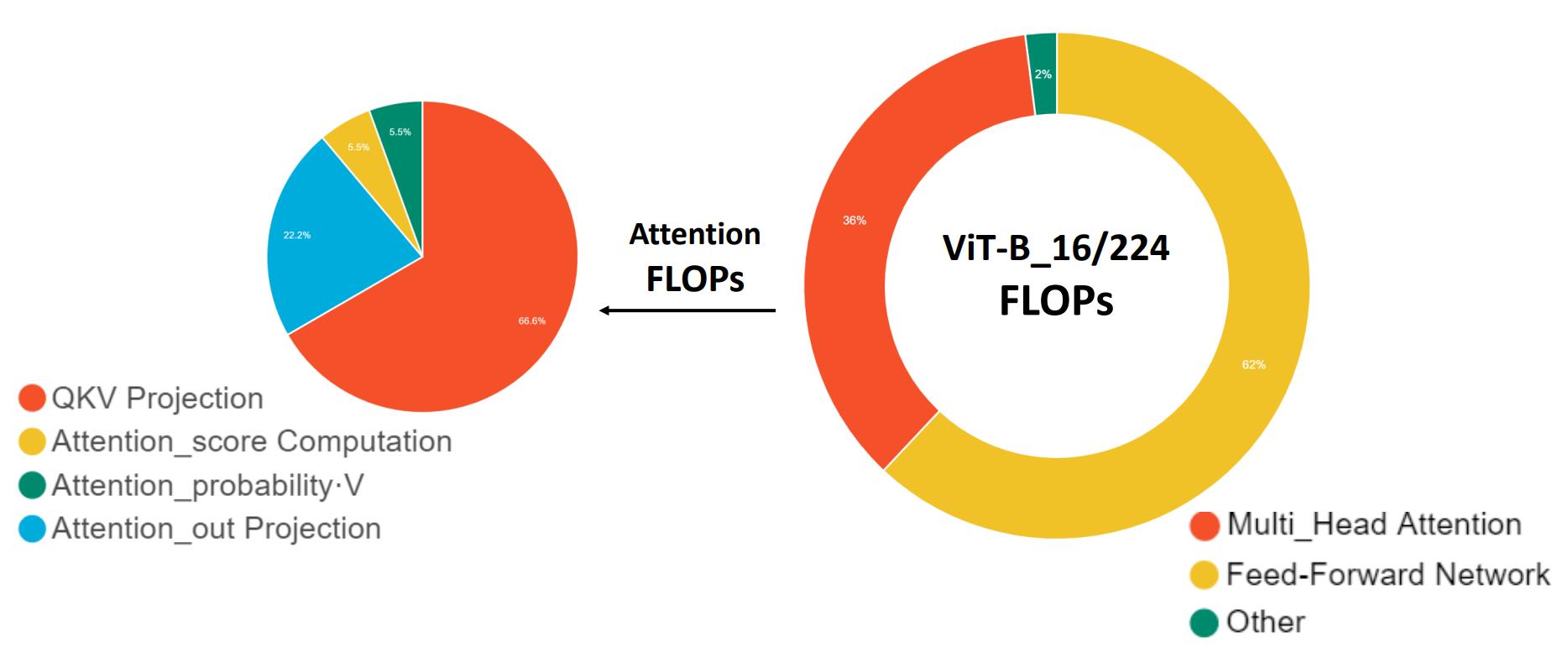}\vspace{-10pt}
\caption{Analysis of FLOPs in ViT model. }
\quad
\label{flops-intro}\vspace{-10pt}
\end{figure}

To achieve better accuracy, ViT demands a significant amount of computation power and memory footprint, which hampers its deployment on mobile devices. 
Typical ViT architecture~\cite{dosovitskiy2020image} includes Multi-Head Self-Attention~(MHSA), Feed-Forward Network~(FFN), Layer Normalization, Activation, and Shortcut modules. 
\cref{flops-intro} shows the proportion of computation amount of each module in ViT-B\_16/224~\cite{dosovitskiy2020image}. It reveals that MHSA and FFN modules occupy the majority of the computations, so this paper mainly reduces the computations of these two modules by pruning redundant patches and heads in a cascade manner. 
Several techniques have been proposed to prune Transformers based on the static distribution of weight values~\cite{han2015deep,zhu2017prune,gordon2020compressing}. 
It appears that patches and heads in ViT are also sparse and have a different impact on accuracy~\cite{chen2021chasing,goyal2020power}. In this paper, we propose patch- and head-based cascade pruning that can dynamically locate informative patches and heads~(collectively called as \textbf{PH-region}), so as to reduce the computation complexity with minimized accuracy loss. Note that, cascade pruning means once an uninformative PH-region is pruned, the involved computations in all following layers will be skipped.


On the other hand, we observe that informative PH-regions appear in previous layers may disappear in the successive layer. 
This is because different layers extract different features and the later layer may ignore informative features extracted by previous layers. 
Therefore, directly executing the cascade pruning only considering uninformative PH-regions in the current layer may destroy the key information captured by previous layers and consequently lead to unacceptable accuracy loss. 
So we define the cumulative score to preserve informative PH-regions across the whole ViT model for better accuracy.

Obviously, setting a uniform pruning ratio for all layers is unacceptable as it may prematurely prune the informative PH-regions in the previous layers. Motivated by the previous work~\cite{wu2020lite}, attention range can represent the capability of extracting inter-dependency between PH-regions in one layer, we smartly leverage the attention range to adjust the pruning ratios for different layers.



In this paper, we propose CP-ViT, a cascade pruning method, to dynamically distinguish and prune uninformative PH-regions. We also define the cumulative score and propose the dynamic pruning ratio adjustment technique to enhance its accuracy and robustness. In this way, the important features will be preserved while the computational amount can be significantly reduced. 
\begin{table}[t]
\renewcommand\arraystretch{1.2}
\centering
\scalebox{0.65}{
\begin{tabular}{|c|c|c|c|c|c|}
\hline
\multicolumn{2}{|c|}{Method}          & CP-ViT & SViTE~\cite{chen2021chasing} & PoWER~\cite{goyal2020power} & VTP~\cite{zhu2021visual} \\ \hline
\multicolumn{2}{|c|}{Dynmaic Pruning} & \checkmark    & \checkmark       & \checkmark     & ×   \\ \hline
\multirow{2}{*}{\begin{tabular}[c]{@{}c@{}}Pruning \\ Granularity\end{tabular}}   & Head-wise & \checkmark & × & × & × \\ \cline{2-6} 
           & Patch-wise               & \checkmark    & ×       & \checkmark     & \checkmark   \\ \hline
\multirow{2}{*}{\begin{tabular}[c]{@{}c@{}}General \\ Applicability\end{tabular}} & Finetune  & \checkmark & \checkmark & \checkmark & \checkmark \\ \cline{2-6} 
           & Without Finetune         & \checkmark    & ×       & ×     & ×   \\ \hline
\end{tabular}
}
\centering
\caption{Comparison of different ViT pruning methods. }\label{method compare}\vspace{-10pt}
\end{table}
Our main contributions can be summarized as follow:
\begin{itemize}
  \item [1)] 
  We explore the sparsity in ViT model and propose a novel cascade pruning method for highly structured PH-regions pruning, which makes the whole process hardware friendly and greatly saves computing resources while maintaining high accuracy.
  \item [2)]
  We define the cumulative score serving the progressive sparsity prediction to identify informative PH-regions on the fly; we also propose to use the maximum value in attention probability to calculate cumulative scores efficiently.
  \item [3)]
  To determine the pruning ratio for each layer, we propose the layer-aware pruning ratio adjustment technique. To the best of our knowledge, this is the first ViT pruning method to dynamically adjust the pruning ratio by leveraging the attention range.
  \item [4)]
  Unlike other ViT pruning methods, CP-ViT can provide superior accuracy with or without finetuning. This greatly enhances CP-ViT's general applicability for practical deployment.
\end{itemize}

To better understanding our CP-ViT, we compare it with other pruning methods, as shown in Table.~\ref{method compare}. Different from VTP~\cite{zhu2021visual}, CP-ViT supports the \textbf{dynamic pruning} for maximally information reservation. Moreover, unlike SViTE~\cite{chen2021chasing}, PoWER~\cite{goyal2020power}, and VTP~\cite{zhu2021visual}, CP-ViT has both \textbf{head-wise} and \textbf{patch-wise} pruning granularity for significant computation saving. Besides, CP-ViT is \textbf{hardware friendly} and has wide \textbf{applicability}.

\section{Related Works}

\subsection{Vision Transformer}
\begin{figure}[tb]
\centering\vspace{-10pt}
\includegraphics[width=0.9\linewidth]{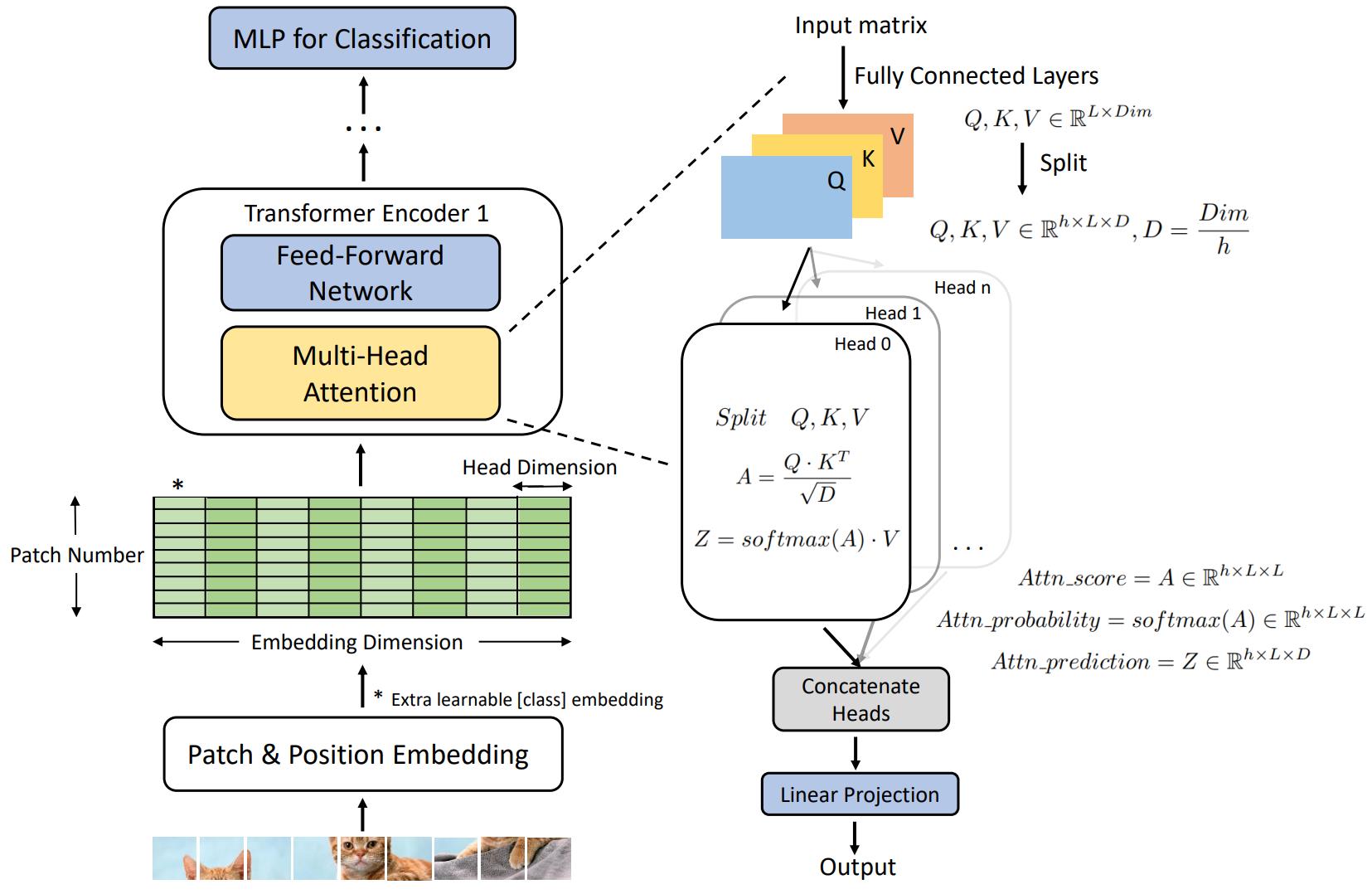}\vspace{-2pt}
\caption{Vision Transformer architecture. On the right is the implementation details of MHSA and definitions of some parameters.}
\label{vit}\vspace{-5pt}
\end{figure}

\begin{figure*}[!htb]
\centering
\includegraphics[width=0.8\linewidth]{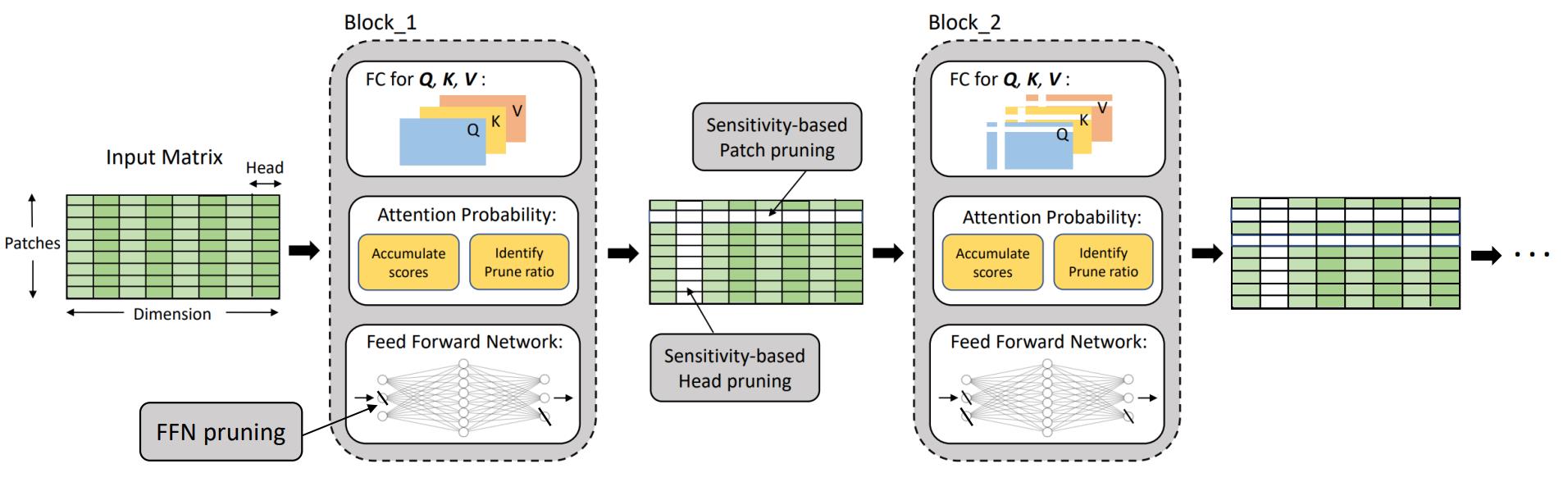}\vspace{-10pt}
\caption{The overall framework of the proposed CP-ViT. Accumulate scores and Identify pruning ratio are two stages inserted behind computation of $attention\_probability$ in MHSA.}
\label{framework}
\end{figure*}

In recent years, Transformer based models~\cite{han2021transformer,wang2021pyramid,touvron2021training} have achieved high accuracy in image classification tasks. Their implementation is similar to Transformers originated from natural language processing, but requires splitting the input images into patches of fixed size, adding ``Classification Token" in each sequence, and feeding the resulting sequence into the standard Transformer Encoder.

Attention of ViT is shown in Fig.~\ref{vit}, it has multiple heads processing a chunk of $Q$, $K$ and $V$ to extract various characteristics between patches and give $attention\_scores$ inside each of heads with the representation of $Q\cdot K^T /\sqrt{D}$. Defining the matrix of $Q\cdot K^T /\sqrt{D}$ as $A$, the dimension of $A$ is $H\times L\times L$, the element $A_h[i,j]$ is the result of inner product between the $i$th row in $Q$ and $j$th column in $K^T$, which can be viewed as the interdependency between $i$th patch and $j$th patch in head $h$. Next, a $softmax$ with row-wise is applied to transform $attention\_scores$ into $attention\_probabilities$. The function of $softmax$ can distinguish the informativeness of patches and enlarge the relative differences between patches with different informativeness. After computing $A\cdot V$, each patch obtains the characteristics from other patches. At the end of attention, the results from multiple heads are merged and reshaped to $L\times dim$ as the output of attention mechanism.

\subsection{Transformer Pruning}\label{transformer pruneing}

ViT has achieved competitive accuracy in a variety of computer vision applications. However, their memory and computing requirements hinder the deployment on mobile devices. MHSA and FFN run extremely slow on mobile devices because their calculation burden is too heavy, which undermines efficiency seriously.



Transformer pruning techniques have been proposed, including weight pruning and conventional token or patch pruning~\cite{zhu2021visual,gordon2020compressing,goyal2020power,chen2021chasing}. Those works have several limitations:

\begin{itemize}
  \item [1)] 
  The weight pruning technique~\cite{gordon2020compressing} statically prune weight and cannot reserve the informative values in the input, which is hard to gain better accuracy.
  \item [2)]
  Conventional input pruning methods often apply the unstructured pruning on input values~\cite{chen2021chasing}, which may achieve better accuracy but is hardware unfriendly.
  \item [3)]
  Most of the existing ViT pruning techniques~\cite{goyal2020power,zhu2021visual} only notice the sparsity in one layer, but without considering the relationship between layers, which may affect the accuracy.
\end{itemize}

Inspired by previous works~\cite{zhu2021visual,goyal2020power,song2020drq}, we further explore the sparsity in patches and heads by the structured cascade pruning. We also define the cumulative score to indicate the informative token or head across layers so as to reserve the real important token or head for high accuracy.


\section{Cascade Vision Transformer Pruning}


To reduce the computational cost and maintain the accuracy, we propose CP-ViT specialized for utilizing the sparsity to prune PH-regions in MHSA and FFN progressively and dynamically. The framework is shown in Fig.~\ref{framework}.

Our study mainly answers the following questions to make the proposed pruning method practical and efficient:

\begin{itemize}
\item 
Whether the sparsity exists during the forward propagation in ViT model, and can we use the sparsity to carry out structured pruning while minimizing the accuracy loss? The answer is yes. Based on our analysis in Section~\ref{algsec1}, we utilize the $attention\_probability$ to prove the existence of sparsity in ViT models. 
\item
How to identify the sparsity in ViT model? In Section~\ref{algsec2}, we define the cumulative score serving progressive sparsity prediction to distinguish informative PH-regions on the fly. Moreover, to efficiently calculate the cumulative score, we also propose to use the maximum value in the $attention\_possibility$.
\item
How to execute the cascade pruning for each layer considering their different sparsity? In Section~\ref{algsec3}, we propose the layer-aware pruning ratio adjustment technique that leverages the attention range to dynamically adjust the pruning ratio of layers. We then implement the cascade pruning on the uninformative PH-regions based on the pruning ratio.
\end{itemize}

\subsection{Informative PH-Regions in ViT}\label{algsec1}

Inspired by the fact that a small number of sensitive regions exist in input feature maps of CNN models~\cite{song2020drq} that have large effect on the accuracy, we try to identify the informative PH-regions in ViT models.
In this section, to study the informativeness of PH-regions in ViT model, we first verify that a small number of key regions in the patches and heads do affect the accuracy and they are termed as \emph{informative PH-regions}, while others are termed as \emph{uninformative PH-regions}.

To identify whether there exist informative PH-regions,
we first calculate the average $attention\_probability$ value of each patch and then divide PH-regions into three segments according to their average magnitudes. This means that segment 1 contains the smallest PH-regions and segment 3 contains the largest PH-regions. We select an input image and the first layer of ViT-B\_16/224 model~\cite{dosovitskiy2020image} as an example to visualize three segments with different luminance, as depicted in Fig.~\ref{seg1}. We then prune each segment one by one and see the Top-1 accuracy results, which are shown in Fig.~\ref{seg2}. It is obvious that segment 3 domains the informative PH-regions, which will lead to significant accuracy loss if we prune the values in it. Alternatively, segments 1 and 2 are less informative, which have a smaller impact on accuracy compared to segment 3.

The above observations indicate that PH-regions with different magnitude have different impacts on the final accuracy. Thus PH-regions with different magnitude can be viewed as PH-regions with different informativeness, which implies applying structured pruning to uninformative PH-regions can accelerate ViT training and inference while maintaining high accuracy.


\begin{figure}[!tb]
\centering\vspace{-5pt}
\includegraphics[width=0.7\linewidth]{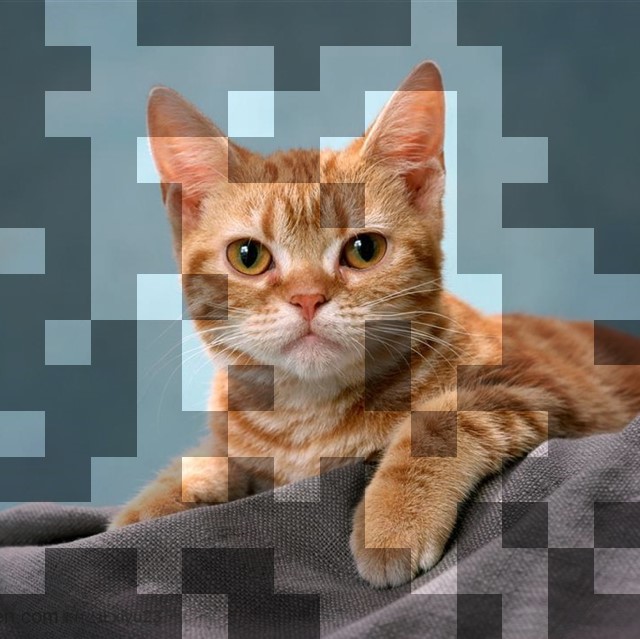}\vspace{-10pt}
\label{dist}\vspace{-10pt}
\end{figure}

\begin{figure}[!tb]
\centering\vspace{-5pt}
\includegraphics[width=0.7\linewidth]{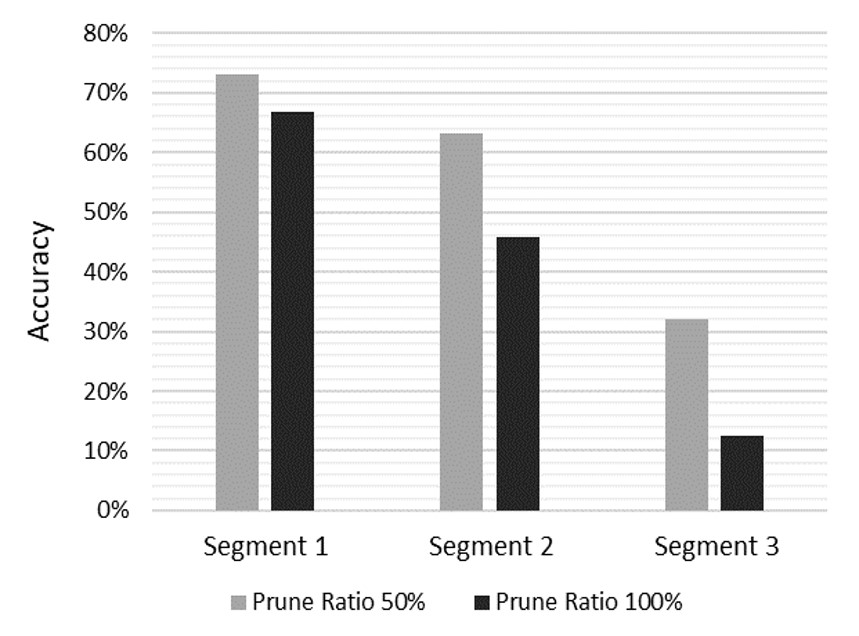}\vspace{-10pt}
\label{dist}\vspace{-10pt}
\end{figure}

\begin{figure}[!tb]
\centering\vspace{-5pt}
\includegraphics[width=0.7\linewidth]{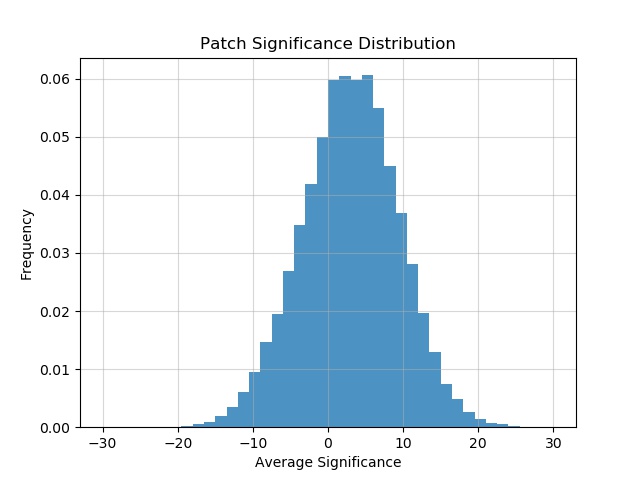}\vspace{-10pt}
\caption{Patch informativeness distribution of ViT-B\_16/224.}
\label{dist}\vspace{-10pt}
\end{figure}

\subsection{Progressive Sparsity Prediction}\label{algsec2}
 
Section~\ref{algsec1} has validated the existence of informative PH-regions. Based on this observation, we need to design an efficient algorithm that can dynamically locate informative PH-regions for sparsity prediction and prune them in FFN and MHSA to accomplish cascade pruning.

We analyse the architecture of MHSA. Consider a Transformer Encoder with attention head $h \in [1,12]$, defining the matrix of $Q\cdot K^T /\sqrt{D}$ as $A$. For a patch $p_0$, $A_h[p_0,:]=\sum_{p}{Q_h[p_0,p]}\centerdot K_h^T[p,:]$ and $A_h[:,p_0]=\sum_{p}{Q_h[:,p]}\centerdot K_h^T[p,p_0]$ represent the interdependency between $p_0$ and other input patches. To be specific, $A_h[p_0,p]$ is computed as the weighted sum of $K_h[p,:]$, which can be viewed as the impact from $p$ to $p_0$ in head $h$. 

The total informativeness of patch $p_0$ in head $h$ can be defined as:
\begin{equation}
    \alpha\sum_{i}{A_h[p_0,i]}+\beta\sum_{j}{A_h[j,p_0]} \label{patch inf1}
\end{equation}
where $\alpha$ and $\beta$ are two parameters indicating the difference between the impact of $p_0$ on other patches and the impact of other patches on $p_0$. Further to say, we can obtain the total informativeness of patch $p_0$ to the whole layer as:
\begin{equation}
    \sum_{h}{(\alpha\sum_{i}{A_h[p_0,i]}+\beta\sum_{j}{A_h[j,p_0]})} \label{patch inf2}
\end{equation}
To define the informativeness of head $h$, the formula is:
\begin{equation}
    \sum_{i}{\sum_{j}{A_h[i,j]}} \label{head inf}
\end{equation}


\begin{algorithm}
\begin{normalsize}
\SetKwInOut{Input}{input}\SetKwInOut{Output}{output}
\caption{MHSA and FFN Pruning}\label{algorithm1}
\Input{
$Q,K,V\in \mathbb{R}^{h\times L\times D}$;\\
Number of heads: $h$;\\
Layer number: $l$;\\
Cumulative patch scores: $s_p\in \mathbb{R}^L$;\\
Cumulative head scores: $s_h\in \mathbb{R}^h$;\\
pruning ratio of last layer: $r_{l-1,p},r_{l-1,h}$;\\
}

\If{$l==0$}{Initialize $l,s_p,s_h$ \\}

 $attention\_probability=softmax(\frac{Q\dot K^T}{\sqrt{D}})$ \\
/* compute the pruning ratio */ \\
$r_{l,p},r_{l,h}$ = Layer-Aware Pruning Ratio \\
/* accumulate informativeness score */ \\
\For{$head=0 \leftarrow h$}{
\For{$patch=0 \leftarrow L$}{
$s_p+=max(attention\_probability[head,:,patch])$ \\
}
$s_h+=sum(s_p)$ \\
}
/* MHSA and FFN pruning */ \\
Threshold\_patch $\epsilon_p=Sort(s_p)[r_{l,p}\cdot L]$ \\
Threshold\_head $\epsilon_h=Sort(s_h)[r_{l,h}\cdot h]$ \\

Generate pruning mask $\Hat{M}_p$ and $\Hat{M}_h$ \\
MHSA and FFN input $\Hat{\mathcal{F}_l} = \mathcal{F}_l\odot \Hat{M}_p\odot \Hat{M}_h$ \\

Update $Normalization$ and $softmax$ \\
\end{normalsize}
\end{algorithm}

After computing the average informativeness of each patch in the 4th layer, we obtain its distribution, as depicted in Fig.~\ref{dist}. The distribution shows that the informativeness of different patches varies greatly. Though Eqn.~\eqref{patch inf1}\eqref{patch inf2}\eqref{head inf} provide rigorous criteria to represent the informativeness of PH-regions, it requires three nested $for$ loops and a huge number of sum operations. This sum-based criteria is inefficient and hinders ViT acceleration. Instead, we need lightweight criteria for representing informativeness.



During forward propagation, $attention\_probability$ is obtained by feeding $attention\_score$ into the $softmax$ function. We observe that numerically similar values in $attention\_score$ may differ by orders of magnitudes in $attention\_probability$. Therefore, the above sum-based criteria can be simplified by directly using the maximum value in $attention\_probability$. Specifically, we obtain the informativeness of the patch by comparing $attention\_probability$ and then choosing the maximum one. Besides, head informativeness is the sum of patch informativeness in this head.

By far, we have obtained informativeness in one layer, but it is not comprehensive to make sparsity prediction only by informativeness in a single layer. As Fig.~\ref{alg2-4} shows, different layers have differences in extracting image features, and the later layers may ignore informative PH-regions captured by previous layers. For example, informative PH-regions in the $2nd$ and $4th$ layers are ignored by $7th$ and $11th$ layers. In $7th$ and $11th$ layers, if we prune these PH-regions only considering the current informativeness, the informative PH-regions captured by $2nd$ and $4th$ layers would be removed in a cascade. This means that the corresponding features extracted by $2nd$ and $4th$ layers will be lost forever, leading to great accuracy loss. 
To minimize accuracy loss, we further define the cumulative scores to represent the informativeness based on the $attention_probability$ of multiple layers rather than a single layer. For each layer, we will accumulate the $attention_probability$ of the current layer and the layers before it so as to obtain the cumulative scores. We sort the cumulative scores and select the smallest $L\times r_{l,p}$ and $H\times r_{l,h}$ scores representing the uninformative patches and heads, where $r_{l,p}$ and $r_{l,h}$ are the pruning ratios of patches and heads respectively.


To perform the cascade pruning, the locations of uninformative PH-regions should be recorded. First, we generate two binary masks $\Hat{M_p},\Hat{M_h}\in \{0,1\}$ locating the uninformative and informative PH-regions~(containing patches and heads). Second, we prune the uninformative PH-regions by conducting element-wise multiplication on the input feature maps and the two masks.
\begin{equation}
    \Hat{\mathcal{F}_l} = \mathcal{F}_l\odot \Hat{M}_p\odot \Hat{M}_h
\end{equation}
As a result, the pruned PH-regions will be set as $0$ while not pruned PH-regions will remain unchanged. 


The detail of the proposed progressive sparsity prediction algorithm is shown in Algorithm.~\ref{algorithm1}. In short, we represent the informativeness in PH-regions by cumulative scores and then prune uninformative PH-regions in a cascade manner. However, setting uniform pruning ratio in all layers is too rough and leads to accuracy degradation. To adjust pruning ratio dynamically, we propose Layer-Aware Cascade Pruning in Section~\ref{algsec3}.

\subsection{Layer-Aware Cascade Pruning}\label{algsec3}

\begin{figure}[t]
\centering\vspace{-5pt}
\includegraphics[width=0.8\linewidth]{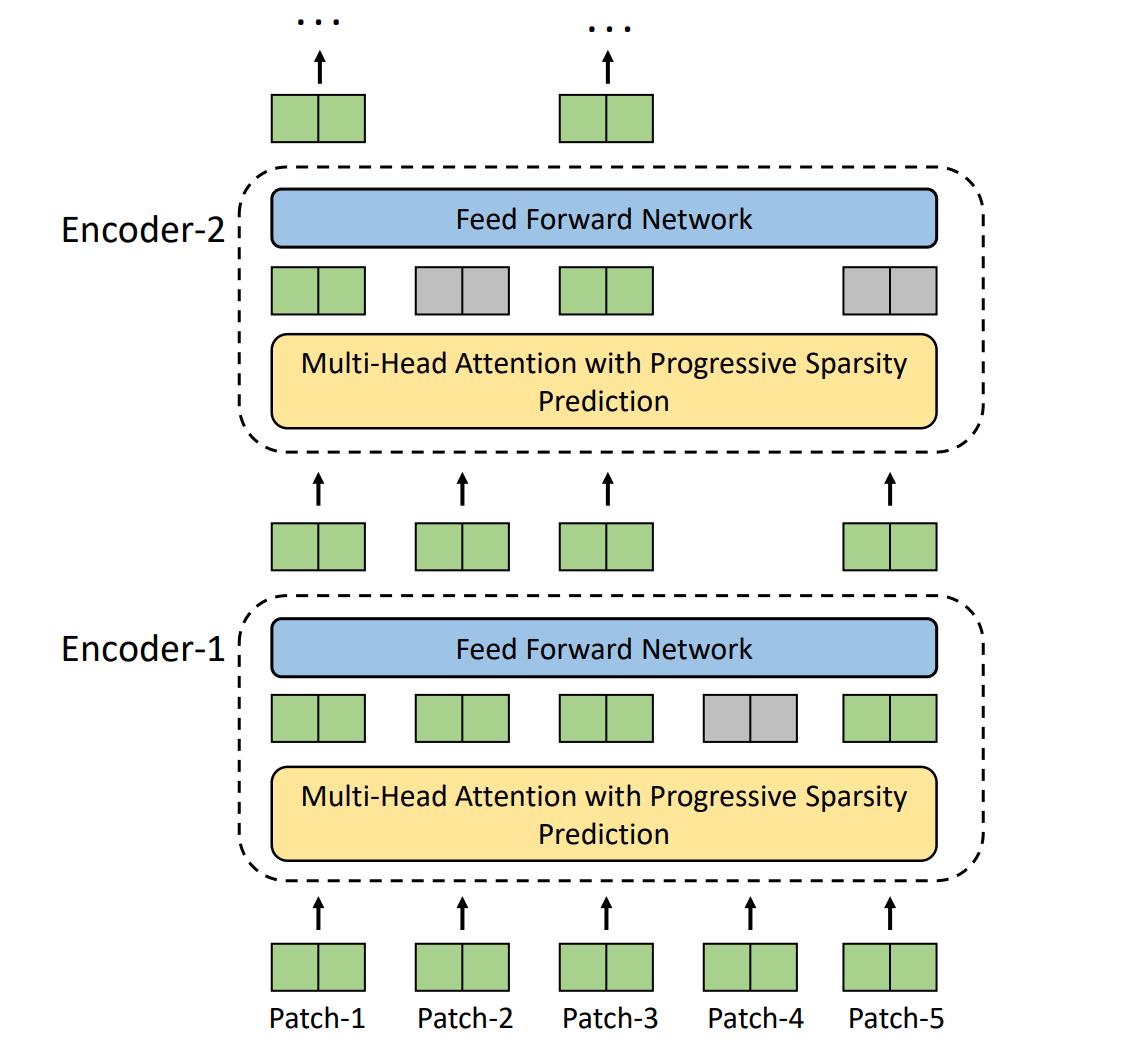}\vspace{-5pt}
\caption{Schematic diagram of cascade pruning, PH-regions in gray indicates being pruned in this layer and involved computations are skipped in following layers.}\vspace{-12pt}
\label{cascade}
\end{figure}

To reduce the computations, the cascade pruning is implemented in our proposed CP-ViT, as depicted in Fig.~\ref{cascade}. But the cascade pruning needs to be controlled precisely according to the characteristics of different layers, otherwise it may prematurely prune the informative PH-regions in the previous layers and lead to accuracy degradation. According to the previous work~\cite{wu2020lite}, attention range can represent the capability of extracting interdependency between PH-regions in one layer. We believe that the short attention range represents the interdependency between PH-regions is unclear so that we cannot leverage $attention\_probability$ to accurately identify the informativeness of PH-regions. On the contrary, the long attention range represents the interdependency between PH-regions is apparent, and $attention\_probability$ can be leveraged to identify the informativeness of PH-regions. In short, we regard the attention range as a guiding role. And we will leverage it to adjust the pruning ratios for different layers. Consequently, we can precisely control the number of pruned PH-regions in each layer to reduces computations while ensuring accuracy.



\begin{figure}[!tb]
\centering
\includegraphics[width=0.8\linewidth]{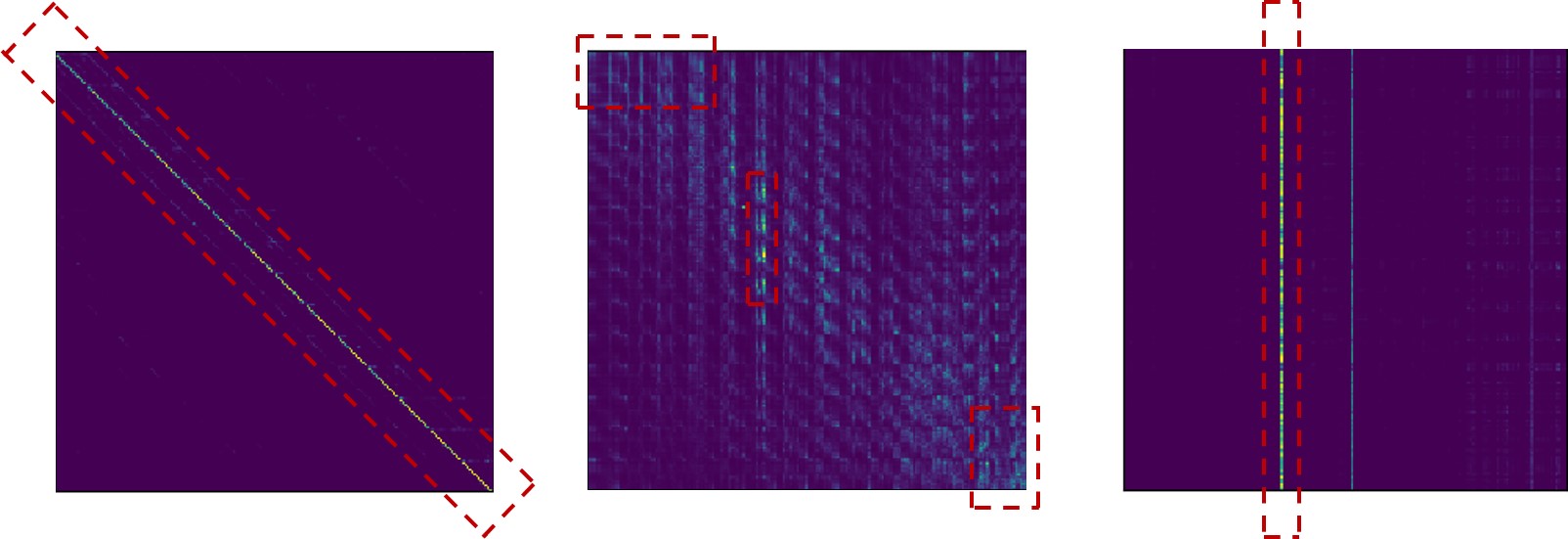}\vspace{-5pt}
\caption{Visualization of $attention\_probability^{L\times L}$ of Head $0$ in $1st$, $6th$, $12th$ layer (from left to right). Bright spots circled in the red box correspond to values that are orders of magnitudes greater than other dark areas.}
\label{head0}\vspace{-12pt}
\end{figure}

The $attention\_probability$ is represented in the dimension of $H\times L\times L$. The element $attention\_probability[h,i,j]$ involves inner product between the $i$th row in $Q$ and $j$th column in $K^T$, which can be viewed as the interdependency between $i$th patch and $j$th patch in head $h$. Fig.~\ref{head0} shows three typical distribution of $attention\_probability$, brighter area means stronger interdependency between corresponding patches:
1) When the bright area is distributed near the diagonal, it indicates that the attention range of this layer is very short, and only the interdependency of patches that are very close to the current patch is captured by this layer. At this time, we cannot identify the informativeness of each patch, so that the pruning ratio should be reduced;
2) When the bright area is distributed on the vertical line, it indicates that the attention range of this layer is very long, and the interdependency of patches that are far from the current patch can also be captured by this layer. We can easily identify the informative patches by locating those bright vertical lines, and we increase the pruning ratio.


\begin{algorithm}
\begin{normalsize}
\SetKwInOut{Input}{input}\SetKwInOut{Output}{output}
\caption{Layer-Aware Pruning Ratio}\label{algorithm2}
\Input{
Attention probability of this layer: $attention\_probability\in \mathbb{R}^{h\times L\times L} $;\\
Attention range offset: $\delta$;\\
Correction factor: $\eta$;\\
Set patch pruning ratio: $r$; \\
pruning ratio of last layer: $r_{l-1}$;\\
}
\If{$l==0$}{Initialize $r_{l-1}=0$ \\}
Generate random ordinates set $S_W=\{s_1,s_2,...,s_k| s_i\in [0,L-1]\}$;\\
\For{each ordinate $s_i$ in $S_W$}{
$s_i'=$max$(attention\_probability[h,s_i,:])$.Col\_index \\
\If{$s_i'\in(s_i-\delta,s_i+\delta)$}{
Short-range attention counter $C_{sr}+=1$ \\
}
}
Attention range $\mathcal{L}=1-\eta\frac{C_{sr}}{k}$ \\
$r_l=r_{l-1}+r\mathcal{L}$ \\
\Output{
Patch pruning ratio of this layer: $r_l$;\\}
\end{normalsize}
\end{algorithm}\label{alg2}

Moreover, to efficiently estimate the attention range and determine pruning ratio, we propose to sample the $attention\_probability$. The detail is shown in Algorithm.~\ref{algorithm2}. Specifically, we first randomly generate $k$ ordinates and store them in $S_W$. Second, for each ordinate value $s_i$ in the $S_W$, we find the abscissa $s_i'$ corresponding to the maximum value in the $s_i$ row of $attention\_probability$. If the difference between $s_i'$ and $s_i$ is less than a pre-defined attention range offset $\delta$, the number of short-attention point~(denote as short-range attention counter) $C_{sr}$ plus one. Third, we obtain the proportion of short-attention point in $k$ points by calculating $\frac{C_{sr}}{k}$. Finally, we obtain the estimated attention range~(denote as $\mathcal{L}$) and the pruning ratio~($r_l$) by calculating $\mathcal{L}=1-\eta\frac{C_{sr}}{k}$ and $r_l=r_{l-1}+r\mathcal{L}$.

\begin{table*}[!t]
\centering
\renewcommand\arraystretch{1.1}  
\scalebox{0.87}{ 

\begin{tabular}{c|c|cc|ccl|ccl}
\hline
\multirow{2}{*}{Model} &
  \multirow{2}{*}{pruning ratio} &
  \multicolumn{2}{c|}{CIFAR-10} &
  \multicolumn{3}{c|}{CIFAR-100} &
  \multicolumn{3}{c}{ImageNet} \\ \cline{3-10} 
 &
   &
  \multicolumn{1}{c}{Acc.(\%)} &
  FLOPs Saving &
  \multicolumn{1}{c}{Acc.(\%)} &
  \multicolumn{2}{l|}{FLOPs Saving} &
  \multicolumn{1}{l}{Top-1 Acc.(\%)} &
  \multicolumn{2}{l}{FLOPs Saving} \\ \hline
\multirow{4}{*}{ViT-B\_16/224~\cite{dosovitskiy2020image}} &
  Baseline~\cite{dosovitskiy2020image} &
  \multicolumn{1}{c}{98.13} &
  - &
  \multicolumn{1}{c}{87.13} &
  \multicolumn{2}{c|}{-} &
  \multicolumn{1}{c}{77.91} &
  \multicolumn{2}{c}{-} \\ \cline{2-10} 
 &
  0.2 &
  \multicolumn{1}{c}{97.77(-0.36)} &
  14.33\% &
  \multicolumn{1}{c}{86.68(-0.45)} &
  \multicolumn{2}{c|}{15.12\%} &
  \multicolumn{1}{c}{77.38(-0.53)} &
  \multicolumn{2}{c}{16.45\%} \\ \cline{2-10} 
 &
  0.3 &
  \multicolumn{1}{c}{97.42(-0.71)} &
  21.54\% &
  \multicolumn{1}{c}{86.32(-0.83)} &
  \multicolumn{2}{c|}{23.87\%} &
  \multicolumn{1}{c}{76.77(-1.14)} &
  \multicolumn{2}{c}{23.02\%} \\ \cline{2-10} 
 &
  0.4 &
  \multicolumn{1}{c}{96.20(-1.93)} &
  29.03\% &
  \multicolumn{1}{c}{84.79(-2.34)} &
  \multicolumn{2}{c|}{32.05\%} &
  \multicolumn{1}{c}{75.09(-2.82)} &
  \multicolumn{2}{c}{32.34\%} \\ \hline
\multirow{4}{*}{ViT-L\_16/224~\cite{dosovitskiy2020image}} &
  Baseline~\cite{dosovitskiy2020image} &
  \multicolumn{1}{c}{97.86} &
  - &
  \multicolumn{1}{c}{86.35} &
  \multicolumn{2}{c|}{-} &
  \multicolumn{1}{c}{76.53} &
  \multicolumn{2}{c}{-} \\ \cline{2-10} 
 &
  0.2 &
  \multicolumn{1}{c}{97.57(-0.29)} &
  15.91\% &
  \multicolumn{1}{c}{86.03(-0.32)} &
  \multicolumn{2}{c|}{16.22\%} &
  \multicolumn{1}{c}{76.03(-0.50)} &
  \multicolumn{2}{c}{17.11\%} \\ \cline{2-10} 
 &
  0.3 &
  \multicolumn{1}{c}{97.14(-0.72)} &
  22.48\% &
  \multicolumn{1}{c}{85.58(-0.77)} &
  \multicolumn{2}{c|}{24.12\%} &
  \multicolumn{1}{c}{75.59(-0.94)} &
  \multicolumn{2}{c}{24.98\%} \\ \cline{2-10} 
 &
  0.4 &
  \multicolumn{1}{c}{96.22(-1.64)} &
  30.11\% &
  \multicolumn{1}{c}{84.57(-1.78)} &
  \multicolumn{2}{c|}{32.44\%} &
  \multicolumn{1}{c}{74.62(-1.91)} &
  \multicolumn{2}{c}{33.01\%} \\ \hline
\multirow{4}{*}{DeiT-B\_16/224~\cite{touvron2021training}} &
  Baseline~\cite{touvron2021training} &
  \multicolumn{1}{c}{99.10} &
  - &
  \multicolumn{1}{c}{90.85} &
  \multicolumn{2}{c|}{-} &
  \multicolumn{1}{c}{81.82} &
  \multicolumn{2}{c}{-} \\ \cline{2-10} 
 &
  0.2 &
  \multicolumn{1}{c}{98.62(-0.48)} &
  16.17\% &
  \multicolumn{1}{c}{90.32(-0.53)} &
  \multicolumn{2}{c|}{14.67\%} &
  \multicolumn{1}{c}{81.06(-0.76)} &
  \multicolumn{2}{c}{15.02\%} \\ \cline{2-10} 
 &
  0.3 &
  \multicolumn{1}{c}{98.36(-0.74)} &
  24.97\% &
  \multicolumn{1}{c}{90.14(-0.71)} &
  \multicolumn{2}{c|}{21.32\%} &
  \multicolumn{1}{c}{80.91(-0.91)} &
  \multicolumn{2}{c}{22.16\%} \\ \cline{2-10} 
 &
  0.4 &
  \multicolumn{1}{c}{98.01(-1.09)} &
  30.08\% &
  \multicolumn{1}{c}{89.68(-1.17)} &
  \multicolumn{2}{c|}{30.92\%} &
  \multicolumn{1}{c}{80.31(-1.51)} &
  \multicolumn{2}{c}{30.67\%} \\ \hline
\end{tabular}

}
\caption{Main results on ImageNet, CIFAR-10, and CIFAR-100 when using CP-ViT without finetuning. We apply our method on three representative ViT models: ViT-B\_16/224~\cite{dosovitskiy2020image}, ViT-L\_16/224~\cite{dosovitskiy2020image} and DeiT-B\_16/224~\cite{touvron2021training}.}\label{exp1}
\end{table*}

\begin{table*}[!tb]
\centering
\renewcommand\arraystretch{1.1}
\scalebox{0.87}{
\begin{tabular}{c|c|cc|ccl|ccl}
\hline
\multirow{2}{*}{Model} &
  \multirow{2}{*}{pruning ratio} &
  \multicolumn{2}{c|}{CIFAR-10} &
  \multicolumn{3}{c|}{CIFAR-100} &
  \multicolumn{3}{c}{ImageNet} \\ \cline{3-10} 
 &
   &
  \multicolumn{1}{c}{Acc.(\%)} &
  FLOPs Saving &
  \multicolumn{1}{c}{Acc.(\%)} &
  \multicolumn{2}{l|}{FLOPs Saving} &
  \multicolumn{1}{l}{Top-1 Acc.(\%)} &
  \multicolumn{2}{l}{FLOPs Saving} \\ \hline
\multirow{4}{*}{ViT-B\_16/224~\cite{dosovitskiy2020image}} &
  Baseline~\cite{dosovitskiy2020image} &
  \multicolumn{1}{c}{98.13} &
  - &
  \multicolumn{1}{c}{87.13} &
  \multicolumn{2}{c|}{-} &
  \multicolumn{1}{c}{77.91} &
  \multicolumn{2}{c}{-} \\ \cline{2-10} 
 &
  0.3 &
  \multicolumn{1}{c}{98.15(+0.02)} &
  22.23\% &
  \multicolumn{1}{c}{87.28(+0.15)} &
  \multicolumn{2}{c|}{24.23\%} &
  \multicolumn{1}{c}{77.75(-0.16)} &
  \multicolumn{2}{c}{24.91\%} \\ \cline{2-10} 
 &
  0.4 &
  \multicolumn{1}{c}{98.02(-0.11)} &
  30.87\% &
  \multicolumn{1}{c}{86.80(-0.33)} &
  \multicolumn{2}{c|}{32.57\%} &
  \multicolumn{1}{c}{77.36(-0.55)} &
  \multicolumn{2}{c}{33.62\%} \\ \cline{2-10} 
 &
  0.5 &
  \multicolumn{1}{c}{97.76(-0.37)} &
  39.43\% &
  \multicolumn{1}{c}{86.16(-0.97)} &
  \multicolumn{2}{c|}{41.03\%} &
  \multicolumn{1}{c}{76.75(-1.16)} &
  \multicolumn{2}{c}{46.34\%} \\ \hline
\multirow{4}{*}{ViT-L\_16/224~\cite{dosovitskiy2020image}} &
  Baseline~\cite{dosovitskiy2020image} &
  \multicolumn{1}{c}{97.86} &
  - &
  \multicolumn{1}{c}{86.35} &
  \multicolumn{2}{c|}{-} &
  \multicolumn{1}{c}{76.53} &
  \multicolumn{2}{c}{-} \\ \cline{2-10} 
 &
  0.3 &
  \multicolumn{1}{c}{97.84(-0.02)} &
  24.17\% &
  \multicolumn{1}{c}{86.39(+0.04)} &
  \multicolumn{2}{c|}{25.37\%} &
  \multicolumn{1}{c}{76.42(-0.11)} &
  \multicolumn{2}{c}{26.21\%} \\ \cline{2-10} 
 &
  0.4 &
  \multicolumn{1}{c}{97.72(-0.14)} &
  31.55\% &
  \multicolumn{1}{c}{86.06(-0.29)} &
  \multicolumn{2}{c|}{34.72\%} &
  \multicolumn{1}{c}{76.21(-0.32)} &
  \multicolumn{2}{c}{34.71\%} \\ \cline{2-10} 
 &
  0.5 &
  \multicolumn{1}{c}{97.52(-0.34)} &
  40.08\% &
  \multicolumn{1}{c}{85.63(-0.72)} &
  \multicolumn{2}{c|}{44.19\%} &
  \multicolumn{1}{c}{75.72(-0.81)} &
  \multicolumn{2}{c}{42.08\%} \\ \hline
\multirow{4}{*}{DeiT-B\_16/224~\cite{touvron2021training}} &
  Baseline~\cite{touvron2021training} &
  \multicolumn{1}{c}{99.10} &
  - &
  \multicolumn{1}{c}{90.85} &
  \multicolumn{2}{c|}{-} &
  \multicolumn{1}{c}{81.82} &
  \multicolumn{2}{c}{-} \\ \cline{2-10} 
 &
  0.3 &
  \multicolumn{1}{c}{99.05(-0.05)} &
  22.15\% &
  \multicolumn{1}{c}{90.98(+0.13)} &
  \multicolumn{2}{c|}{24.82\%} &
  \multicolumn{1}{c}{81.66(-0.16)} &
  \multicolumn{2}{c}{22.62\%} \\ \cline{2-10} 
 &
  0.4 &
  \multicolumn{1}{c}{98.84(-0.26)} &
  34.12\% &
  \multicolumn{1}{c}{90.76(-0.09)} &
  \multicolumn{2}{c|}{32.46\%} &
  \multicolumn{1}{c}{81.52(-0.30)} &
  \multicolumn{2}{c}{32.41\%} \\ \cline{2-10} 
 &
  0.5 &
  \multicolumn{1}{c}{98.42(-0.68)} &
  39.09\% &
  \multicolumn{1}{c}{90.37(-0.48)} &
  \multicolumn{2}{c|}{43.02\%} &
  \multicolumn{1}{c}{81.13(-0.69)} &
  \multicolumn{2}{c}{41.62\%} \\ \hline
\end{tabular}
}
\caption{Main results on ImageNet, CIFAR-10, and CIFAR-100 when using CP-ViT with finetuning.}\label{exp2-1}\vspace{-10pt}
\end{table*}

\section{Experiment}

In this section, we will demonstrate the effectiveness and general applicability of CP-ViT through extensive experiments. We use ImageNet~\cite{deng2009imagenet}, CIFAR-10~\cite{krizhevsky2009learning}, and CIFAR-100~\cite{krizhevsky2009learning} to verify our method. For a fair comparison, we utilize the official implementations of ViT models and the accuracy results released in their paper~\cite{dosovitskiy2020image,touvron2021training}. 
To prove the applicability on a wide range of models based on Vision Transformer, we apply CP-ViT to ViT~\cite{dosovitskiy2020image} and DeiT~\cite{touvron2021training} models with different parameter scales from 5.7M to 307.4M.
We directly apply CP-ViT to pre-trained models and report the results in Section~\ref{expsec1} to prove the accuracy without finetuning. Moreover, we finetune the models and report the results in Section~\ref{expsec2} to validate the effectiveness of CP-ViT with finetuning.
We also compare CP-ViT with other ViT pruning methods~\cite{zhu2021visual,goyal2020power,chen2021chasing} on ImageNet dataset in Section~\ref{expsec3}.

For implementation details, we finetune the model for 20 epochs using SGD with a start learning rate of 0.02 and cosine learning rate decay strategy on CIFAR-10 and CIFAR-100; we also finetune on ImageNet for 30 epochs using SGD with a start learning rate of 0.01 and weight decay 0.0001. All codes are implemented in PyTorch, and the experiments are conducted on 2 Nvidia Volta V100 GPUs.

\subsection{CP-ViT without Finetuning}\label{expsec1}

In this section, we apply CP-ViT to ViTs and DeiTs without finetuning, which is a challenging task because CP-ViT must match the weights of pre-trained models. In Table.~\ref{exp1}, we present the results on CIFAR-10, CIFAR-100, and ImageNet with a range of pruning ratio from 0.2 to 0.4. From these results, we observe that our method achieves 25\% drop in floating point operations~(FLOPs) without finetuning while the accuracy loss is less than 1\%. Even when reducing 30\% to 40\% FLOPs, the accuracy declines by no more than 3\%. As the complexity of CIFAR-10/100 is lower than that of ImageNet, CP-ViT achieves higher accuracy on CIFAR-10 and CIFAR-100. Since the scale of ViT-L\_16/224 is larger than that of ViT-B\_16/224, the accuracy loss of ViT-L\_16/224 is lower than that of ViT-B\_16/224. 

\begin{table*}[!tb]
\centering
\renewcommand\arraystretch{1.0}
\scalebox{0.87}{
\begin{tabular}{c|c|cc|cc}
\hline
\multirow{2}{*}{Model} & \multirow{2}{*}{Method} & \multicolumn{2}{c|}{Not Finetune} & \multicolumn{2}{c}{Finetune} \\ \cline{3-6} 
                                &              & \multicolumn{1}{c}{Top-1 Acc.(\%)} & FLOPs Saving & \multicolumn{1}{c}{Top-1 Acc.(\%)} & FLOPs Saving \\ \hline
\multirow{5}{*}{DeiT-Ti\_16/224~\cite{touvron2021training}} & Baseline~\cite{touvron2021training}                & \multicolumn{1}{c}{72.20}   & -  & \multicolumn{1}{c}{72.20} & - \\ \cline{2-6} 
                                & VTP~\cite{zhu2021visual}          & \multicolumn{1}{c}{69.37(-2.83)}   & 21.68\%      & \multicolumn{1}{c}{70.55(-1.65)}  & 45.32\%      \\ \cline{2-6} 
                                & PoWER~\cite{goyal2020power}        & \multicolumn{1}{c}{69.56(-2.64)}   & 20.32\%      & \multicolumn{1}{c}{70.05(-2.15)}  & 41.26\%      \\ \cline{2-6} 
                                & HVT~\cite{pan2021scalable}          & \multicolumn{1}{c}{68.43(-3.77)}   & 21.17\%      & \multicolumn{1}{c}{70.01(-2.19)}  & 47.32\%      \\ \cline{2-6} 
                                & CP-ViT(Ours) & \multicolumn{1}{c}{71.06(-1.14)}   & 23.02\%      & \multicolumn{1}{c}{71.24(-0.96)}  & 43.34\%      \\ \hline
\multirow{5}{*}{DeiT-S\_16/224~\cite{touvron2021training}} & Baseline~\cite{touvron2021training}     & \multicolumn{1}{c}{79.80}          & -            & \multicolumn{1}{c}{79.80}         & -            \\ \cline{2-6} 
                                & VTP~\cite{zhu2021visual}          & \multicolumn{1}{c}{77.35(-2.45)}   & 20.74\%      & \multicolumn{1}{c}{78.24(-1.56)}  & 42.52\%      \\ \cline{2-6} 
                                & PoWER~\cite{goyal2020power}        & \multicolumn{1}{c}{77.02(-2.78)}   & 21.46\%      & \multicolumn{1}{c}{78.30(-1.50)}  & 41.36\%      \\ \cline{2-6} 
                                & HVT~\cite{pan2021scalable}          & \multicolumn{1}{c}{76.72(-3.08)}   & 20.52\%      & \multicolumn{1}{c}{78.05(-1.75)}  & 47.80\%      \\ \cline{2-6} 
                                & CP-ViT(Ours) & \multicolumn{1}{c}{78.84(-0.96)}   & 20.96\%      & \multicolumn{1}{c}{79.08(-0.72)}  & 42.24\%      \\ \hline
\multirow{5}{*}{DeiT-B\_16/224~\cite{touvron2021training}} & Baseline~\cite{touvron2021training}     & \multicolumn{1}{c}{81.82}          & -            & \multicolumn{1}{c}{81.82}         & -            \\ \cline{2-6} 
                                & VTP~\cite{zhu2021visual}          & \multicolumn{1}{c}{79.46(-2.36)}   & 19.84\%      & \multicolumn{1}{c}{80.70(-1.12)}  & 43.20\%      \\ \cline{2-6} 
                                & PoWER~\cite{goyal2020power}        & \multicolumn{1}{c}{79.09(-2.73)}   & 20.75\%      & \multicolumn{1}{c}{80.17(-1.65)}  & 39.24\%      \\ \cline{2-6} 
                                & HVT~\cite{pan2021scalable}          & \multicolumn{1}{c}{78.88(-2.94)}   & 20.14\%      & \multicolumn{1}{c}{79.94(-1.88)}  & 44.78\%      \\ \cline{2-6} 
                                & CP-ViT(Ours) & \multicolumn{1}{c}{80.91(-0.91)}   & 22.16\%      & \multicolumn{1}{c}{81.13(-0.69)}  & 41.62\%      \\ \hline
\end{tabular}
}
\caption{Comparison with different ViT pruning methods on ImageNet dataset. The accuracy without finetuning was not mentioned in VTP~\cite{zhu2021visual}, PoWER~\cite{goyal2020power} and HVT~\cite{pan2021scalable}, we replicate their code and obtain the accuracy without finetuning. }\label{exp2-2}\vspace{-10pt}
\end{table*}

\subsection{Finetuning CP-ViT models}\label{expsec2}

To achieve higher accuracy, we finetune ViT models on ImageNet, CIFAR-10, and CIFAR-100. The results are summarized in Table.~\ref{exp2-1}.

Compared with the baseline, CP-ViT achieves nearly 2$\times$ FLOPs reduction while maintaining the accuracy loss within 1\%, which indicates that CP-ViT can significantly reduce computational redundancy while reserving accuracy. Moreover, compared with Table.~\ref{exp1}, finetuning CP-ViT greatly increases accuracy and enables it to achieve more FLOPs reduction in a larger pruning ratio. As in Table.~\ref{exp2-1}, it is also impressive that when the pruning ratio is 0.3, CP-ViT not only achieves over 20\% FLOPs reduction but also surpasses the baseline by around 0.1\% accuracy, which means finetuning CP-ViT improves the ability to extract informative input features and make accurate image recognition.


\subsection{Comparison on ImageNet}\label{expsec3}

In this section, we compare the CP-ViT with three state-of-the-art ViT pruning methods. Since CP-ViT can be deployed with different pruning ratios, we select an ideal pruning ratio and then compare CP-ViT method with other ViT pruning methods, as in Table.~\ref{exp2-2}.

CP-ViT exhibits favorable trade-offs between accuracy and efficiency whether with or without finetuning. Compared with other pruning methods, the accuracy of CP-ViT improved by 2\% when FLOPs is reduced by the same amount without finetuning. Moreover, it is notable that CP-ViT progressively reduces over 40\% FLOPs while making accuracy loss within 1\% when finetuning pre-trained models, while other methods still have up to 2.2\% accuracy loss. In summary, compared to other pruning techniques, CP-ViT is more efficient with higher accuracy.

\subsection{Ablation Study}

We conduct ablation study on ImageNet dataset to validate the proposed CP-ViT. We tune the pruning ratio from 0.2 to 0.5.

\begin{figure}[tb]
\centering\vspace{-5pt}
\includegraphics[width=0.8\linewidth]{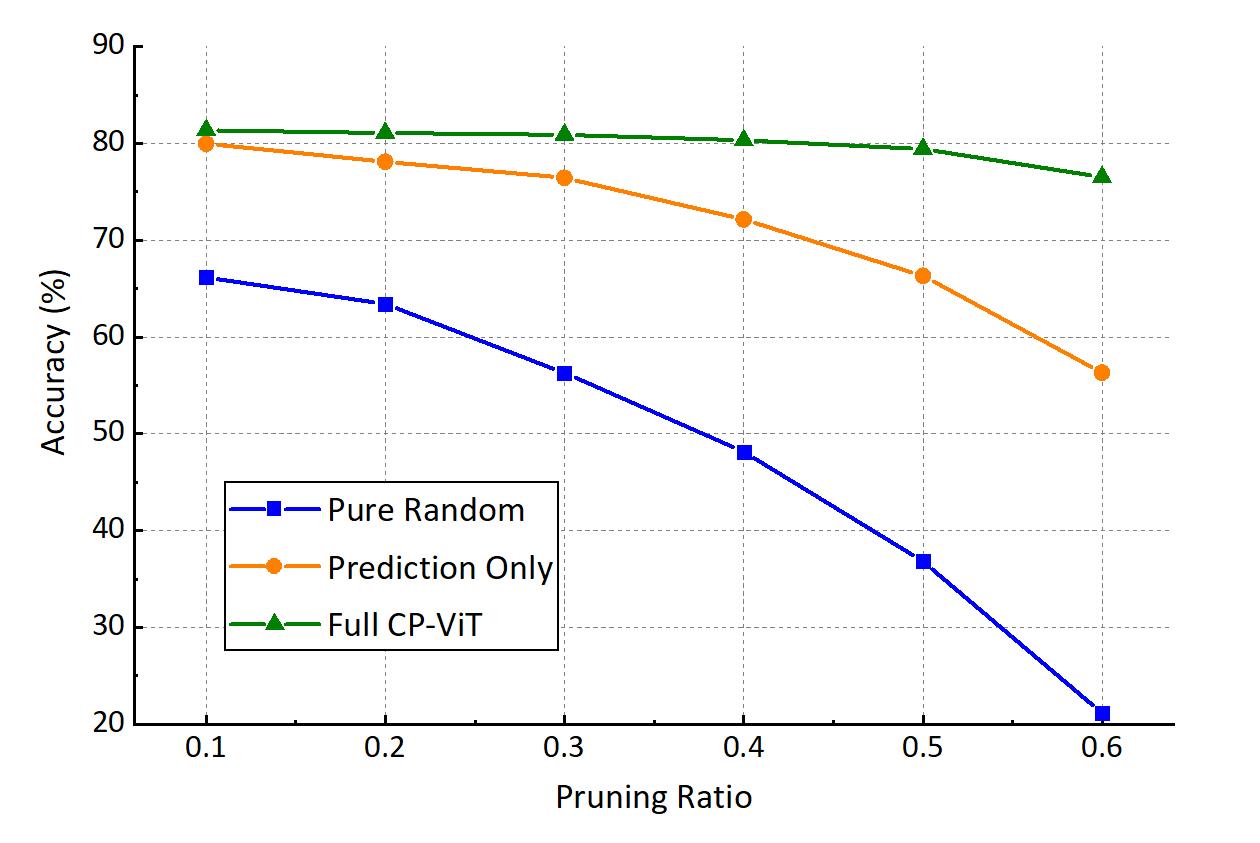}\vspace{-10pt}
\caption{Comparison between different pruning methods.}
\quad
\label{ablation2}\vspace{-12pt}
\end{figure}

The Progressive Sparsity Prediction and the Layer-Aware Cascade Pruning are indispensable to the accuracy of ViT. To validate their effectiveness, we set up the experiment in a progressive manner: 1) we randomly prune patches (denote as Pure Random) to verify the effectiveness of Progressive Sparsity Prediction; 2) we apply Progressive Sparsity Prediction to locate the uninformative PH-regions but use a uniform pruning ratio for all layers (denote as Prediction Only) to verify the effectiveness of Layer-Aware Cascade Pruning; 3) we apply both Progressive Sparsity Prediction and Layer-Aware Cascade Pruning (denote as CP-ViT). The results are shown in Table.~\ref{ablation} and Fig.~\ref{ablation2}. And we can find that CP-ViT achieves higher accuracy compared to Pure Random and Prediction Only. 

\begin{table}[!tb]
\centering
\renewcommand\arraystretch{1.1}
\scalebox{0.75}{
\begin{tabular}{l|c|ccc}
\hline
\multicolumn{1}{c|}{\multirow{2}{*}{Model}} & \multirow{2}{*}{Prune Ratio} & \multicolumn{3}{c}{Top-1 Acc.(\%)}                                                   \\ \cline{3-5} 
\multicolumn{1}{c|}{}                       &                              & \multicolumn{1}{c}{Pure Random} & \multicolumn{1}{c}{Prediction Only} & CP-ViT \\ \hline
\multirow{6}{*}{DeiT-B\_16/224}              & 0.1                          & \multicolumn{1}{c}{66.17}       & \multicolumn{1}{c}{79.96}           & 81.37       \\ \cline{2-5} 
                                             & 0.2                          & \multicolumn{1}{c}{63.41}       & \multicolumn{1}{c}{78.12}           & 81.06       \\ \cline{2-5} 
                                             & 0.3                          & \multicolumn{1}{c}{56.28}       & \multicolumn{1}{c}{76.44}           & 80.91       \\ \cline{2-5} 
                                             & 0.4                          & \multicolumn{1}{c}{48.16}       & \multicolumn{1}{c}{72.15}           & 80.31       \\ \cline{2-5} 
                                             & 0.5                          & \multicolumn{1}{c}{36.82}       & \multicolumn{1}{c}{66.31}           & 79.42       \\ \cline{2-5} 
                                             & 0.6                          & \multicolumn{1}{c}{21.15}       & \multicolumn{1}{c}{56.32}           & 76.53       \\ \hline
\end{tabular}
}
\caption{Main results when applying different pruning methods. }\label{ablation}\vspace{-5pt}
\end{table}

By comparing Pure Random and CP-ViT, we can conclude that Progressive Sparsity Prediction can preserve informative PH-regions and therefore is necessary for achieving accurate ViT. Moreover, as shown in Fig.~\ref{ablation2}, Prediction Only can reserve the accuracy when the pruning ratio is small, but it cannot maintain the accuracy at the large pruning ratio. This indicates that the Layer-Aware Cascade Pruning is useful especially at large pruning ratio. In summary, assisted by Progressive Sparsity Prediction and Layer-Aware Cascade pruning techniques, CP-ViT can guarantee accuracy even with large pruning ratio.

\section{Conclusion}

In this paper, we propose Cascade Vision Transformer Pruning via Progressive Sparsity Prediction, termed CP-ViT, to dynamically locate uninformative patches and heads, and conduct structured pruning on them for reducing computations. We assign cumulative scores for input patches and heads according to their maximum value in $attention probability$, which greatly simplifies the sparsity prediction. We also propose layer-aware cascade pruning that can dynamically adjust pruning ratio for each layer based on the attention range. Our evaluation shows that CP-ViT scheme outperforms other similar schemes in performance and accuracy. 


{\small
\bibliographystyle{ieee_fullname}
\bibliography{egbib}
}

\end{document}